\documentclass[letterpaper]{article}
\usepackage[preprint]{aaai2027}

\usepackage[hyphens]{url}  
\usepackage{graphicx} 
\usepackage{natbib}  
\usepackage{caption} 
\usepackage{algorithm}
\usepackage{algorithmic}
\usepackage{amsmath}
\usepackage{amssymb}
\usepackage{booktabs}
\usepackage[table]{xcolor}
\usepackage{colortbl}
\usepackage{multirow}
\usepackage{pifont}
\definecolor{oursgreen}{RGB}{215,242,220}
\definecolor{poscolor}{RGB}{0,150,0}   
\definecolor{negcolor}{RGB}{200,0,0}   

\title{
    FoMoVLA: Bridging Visual Foresight and Motion Guidance for Vision-Language-Action Models
}
\author{
    Wei Li$^{1,*}$,
    Peijin Jia$^{1,*}$,
    Yuan Ma$^{1,*,\ddagger}$,
    Xuefeng Jiang$^{2}$,
    Titong Jiang$^{3}$,
    Sheng Sun$^{2}$,\\
    Yujian Li$^{1}$,
    Xin Wen$^{1}$,
    Han Hong$^{1}$,
    Zhikang Liu$^{1}$,
    Bailin Li$^{1}$,
    Kun Zhan$^{1,\dagger}$
}
\affiliations{
    $^1$~LiAuto \quad
    $^2$~University of Chinese Academy of Sciences \quad
    $^3$~Tsinghua University\\
    $^*$~Equal contribution.\quad
    $^\ddagger$~Project lead.\quad
    $^\dagger$~Corresponding author.
}

\begin{document}

\maketitle

\begin{abstract}
Vision-Language-Action (VLA) models have achieved impressive results in visuomotor policy learning, yet remain fundamentally reactive, mapping current observations and language to actions without explicit forward prediction of world dynamics.
Existing visual foresight methods predict future visual states but lack explicit motion guidance — they show where to go but not how to get there. We argue that future feature prediction and sparse point tracking are naturally complementary: the former provides the goal state, while the latter captures the continuous motion path toward it.
We propose
\textbf{FoMoVLA}, 
a framework that augments VLA representations with explicit spatio-temporal supervision by jointly learning future feature foresight and sparse 2D point tracking, enhancing the continuous action policy. FoMoVLA introduces compact foresight tokens to decode future feature states, decodes sparse temporal 2D point trajectories to model compact geometric motion, and couples both through a lightweight future-conditioned cross-attention module that enables consistent reasoning between anticipated states and point dynamics.
Extensive experiments on LIBERO, RoboCasa GR-1 Tabletop, and LIBERO-Plus demonstrate the state-of-the-art performance and strong zero-shot generalization. Project page is available at \url{https://liauto-research.github.io/FoMoVLA}.
\end{abstract}

\section{Introduction}
Recent advances in robot learning have established VLA models as a powerful paradigm for robotic manipulation, demonstrating strong performance across diverse manipulation tasks~\cite{brohan2023rt2,kim2024openvla,li2024cogact,team2024octo,black2024pi0,nvidia2025groot}. However, 
existing VLA approaches typically rely solely on instantaneous visual observations and language instructions to learn direct reactive mappings. While effective in canonical embodied tasks, this reactive paradigm inherently lacks the spatio-temporal foresight required to model future scene evolution or long-horizon object dynamics.

\begin{figure}[t]
    \centering
    \includegraphics[width=\columnwidth]{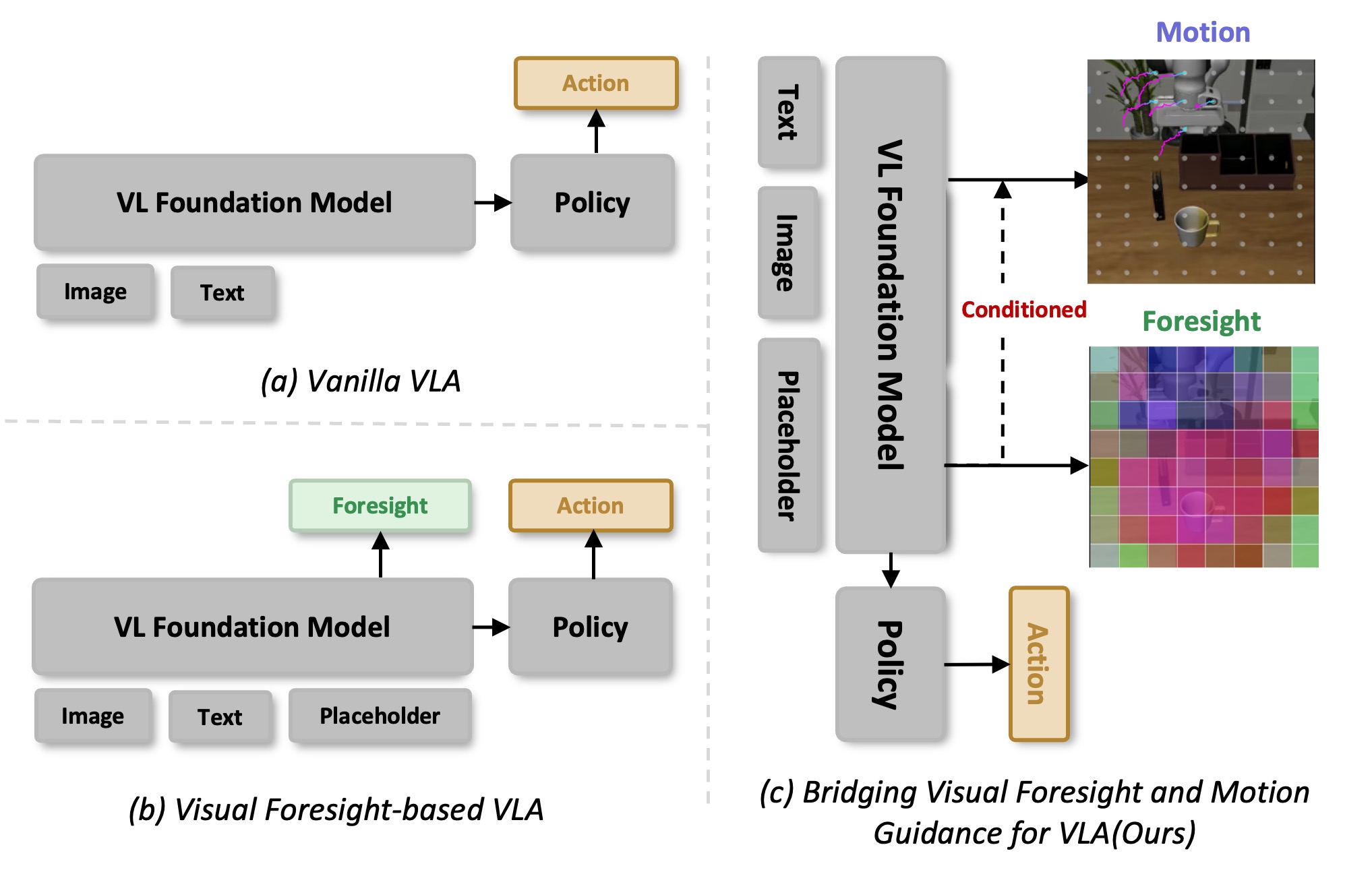}
     \caption{Architecture comparison. (a) Vanilla VLA directly maps observations to actions. (b) VLA augmented with visual foresight as an auxiliary task. (c) FoMoVLA jointly predicts visual foresight and motion, enforcing their consistency to significantly enhance action generation.}
    \label{fig:intro_comparison}
\end{figure}

A growing research line seeks to enhance the foresight capability of VLA models through future-predictive supervision \cite{zhang2025dreamvla,bu2025univla,lara2026,HiF-VLA}. However, existing approaches still fall short when it comes to truly serving action control. On the one hand, dense pixel-level prediction of multi-frame future visual states preserves rich scene information but inevitably captures large amounts of static content irrelevant to control~\cite{zhang2026anticipation}, leading to representational redundancy and high modeling overhead. On the other hand, more compact keyframe-level future representations improve efficiency yet typically only characterize the target state to be reached, failing to capture the concrete dynamic interactions required to arrive at that state~\cite{zhong2025flowvla,HiF-VLA}. 

We argue that actionable foresight should encode not only the future state itself but also the continuous temporal motion dynamics leading toward it. Building on this insight, we realize visual foresight by predicting future features instead of raw pixels and motion guidance by sparse point tracking that encodes where task-relevant objects will move.
Beyond merely predicting future visual pixels, motion tracking~\cite{karaev2024cotracker, zhang2026molmomotion, qian2026geopredict} via 2D point trajectories offers a more action-aligned spatio-temporal representation of world dynamics, capturing the continuous local movements and interaction details throughout task execution. Crucially, we use future feature as a goal-state anchor and unify terminal-state modeling with temporal point tracking to jointly guide action generation. In this way, future imagination is no longer an isolated auxiliary objective. Instead, it works alongside point tracking to channel foresight directly into action generation, enabling the policy to learn actions that are both future-oriented and physically constrained.

To this end, we propose \textbf{FoMoVLA}, which jointly learns future feature prediction and sparse 2D point tracking as complementary training-only supervision within a foundation model. Specifically, several learnable foresight tokens are supervised to learn the visual foresight via an EMA teacher, providing a compact goal-state representation, while a frozen off-the-shelf point tracker provides per-frame displacement targets for sparse grid points, forcing image-token hidden states to encode explicit motion cues. Crucially, the two objectives are not independent: a zero-initialized cross-attention module (FCCA) conditions point tracking on the foresight representation, anchoring local trajectory estimation to a global foresight prior. At inference, all auxiliary branches are discarded, incurring zero parameter overhead and negligible latency.
Our main contributions are as follows:
\begin{itemize}
    \item We introduce a novel VLA framework that augments latent representations with explicit spatio-temporal supervision by jointly learning compact future feature prediction and sparse 2D point tracking.
    \item 
    We propose a future-conditioned cross-attention module that couples foresight tokens with temporal point tracking, ensuring consistency between anticipated visual states and geometric motion dynamics.
    \item We achieve the state-of-the-art performance on LIBERO and RoboCasa GR-1 Tabletop, along with strong zero-shot generalization on LIBERO-Plus, outperforming competitive VLA baselines across diverse manipulation tasks.
\end{itemize}

\section{Related Work}

\paragraph{Vision-Language-Action Models.}
By directly fine-tuning pretrained vision-language models (VLMs) on robot demonstration data, Vision-Language-Action (VLA) models map visual observations and language instructions to robot actions.~\cite{brohan2023rt2,kim2024openvla,team2024octo}. The field has since advanced along two primary axes. On the policy side, more expressive action decoders have been developed, including flow-matching~\cite{black2024pi0,pi2026pi07} and diffusion-based formulations~\cite{chi2023diffusion,liu2025rdt,li2024cogact}. On the perception side, stronger backbones with dual-system architectures~\cite{nvidia2025groot} and specialized vision encoders~\cite{shi2026memoryvla} have been adopted to improve visual grounding. Despite these advances, recent studies indicate that the primary bottleneck in embodied control lies in the visual module of VLMs, rather than in their language understanding or action decoding capabilities~\cite{zhang2026vlm4vla}. This suggests that improvements in downstream manipulation require not merely scaling the VLM backbone, but fundamentally enhancing its ability to represent visual structure and dynamics.

\paragraph{Predictive Supervision and Motion Tracking.}
To bridge this gap, a growing line of work incorporates future prediction into VLA training through explicit or implicit predictive objectives. Pixel-level approaches forecast future frames or interleave visual generation with action prediction~\cite{cen2025worldvla,zhang2025dreamvla,hu2026bagelvla,cai2026internvla}, providing rich supervision at the cost of substantial computational overhead. Latent-space methods offer a more efficient alternative by predicting future representations directly in the VLM feature space~\cite{sun2026vlajepa,fan2026futurevla,lyu2026lda,su2026wog,ma2025vita,syed2026ahead}. Geometric approaches capture motion more compactly: FlowVLA predicts pixel-level optical flow as a visual chain-of-thought~\cite{zhong2025flowvla}, JOPAT jointly denoises sparse 2D point tracks, visibility masks, and actions within a diffusion framework~\cite{guan2026jopat}, and Spatial Forcing distills geometric knowledge from a frozen teacher as auxiliary supervision during training~\cite{li2026spatialforcing}. These methods indicate that motion and geometric information can enhance action prediction. However, these methods learn visual foresight or motion tracking separately, without explicit interaction. In contrast, our approach jointly learns future feature prediction and point tracking within a shared backbone.

\section{Method}

\begin{figure*}[t]
    \centering
    \includegraphics[width=1.0\textwidth]{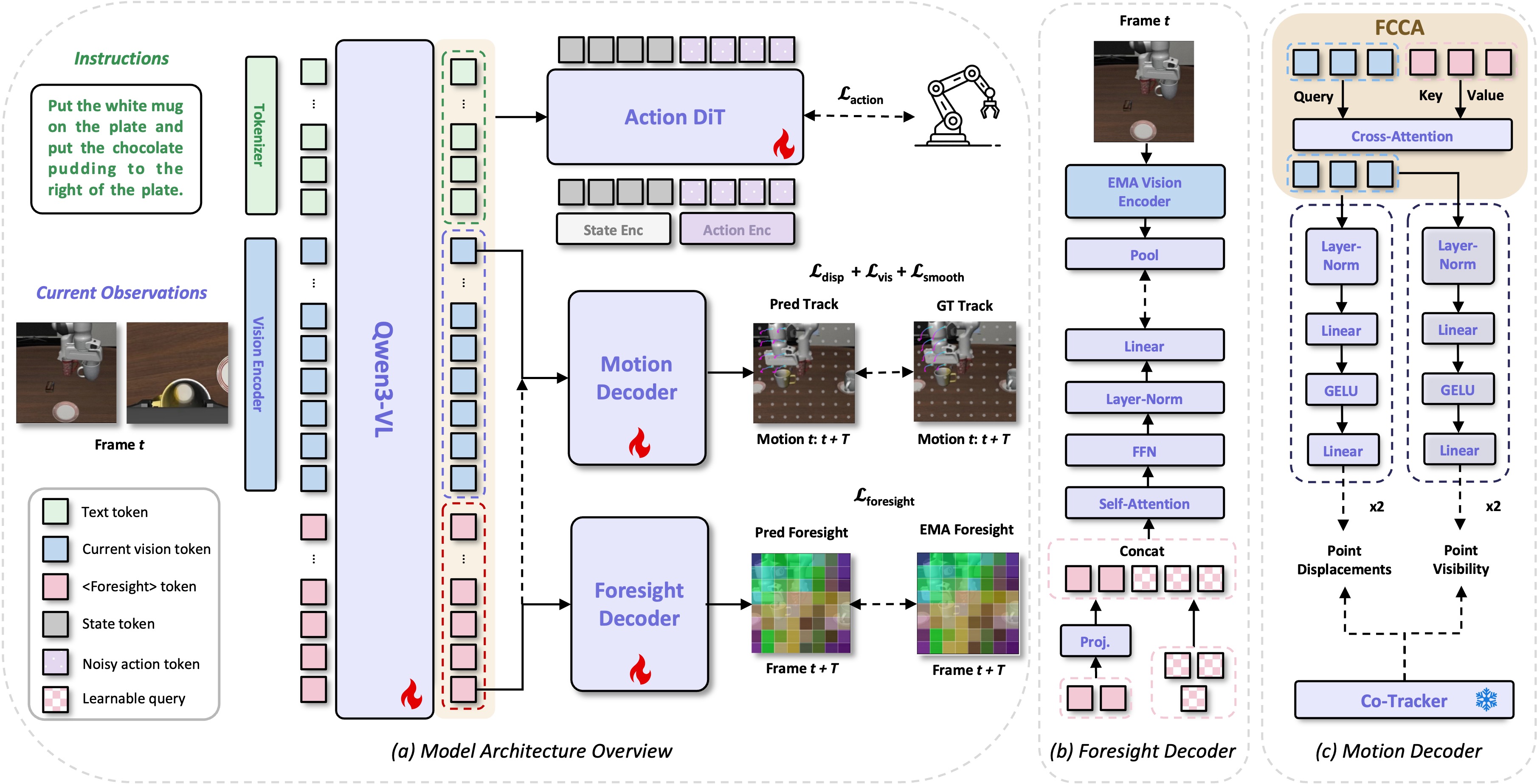}
    \caption{Overview of FoMoVLA. Given a current observation and language instruction, the VLM backbone processes image tokens (with text tokens placed first for goal-aware conditioning) alongside $K$ learnable \texttt{<Foresight>} tokens.}
    \label{fig:main_method}
\end{figure*}

Our framework, illustrated in Fig.~\ref{fig:main_method}, augments a VLA policy with three complementary training-time branches that inject explicit spatio-temporal supervision into the learned representations. Specifically, it comprises: (1) a point tracking branch that predicts 2D point trajectories from image token hidden states under the supervision of a frozen point tracker, grounding image features in explicit motion cues (\emph{how} to move) (Section~\ref{sec:cotracker}); (2) a future feature prediction branch to learn a compact representation of the future visual state (\emph{where} to end up) (Section~\ref{sec:foresight}); and (3) a future-conditioned cross-attention (FCCA) module that couples these two objectives by conditioning motion prediction on the predicted future (Section~\ref{sec:goal_cross_attn}). 

\subsection{Preliminaries}
\label{sec:prelim}

\paragraph{VLA Policy Formulation.}
A typical VLA model defines a policy $\pi$ that maps a visual observation $o_t \in \mathbb{R}^{H \times W \times 3}$ and a natural-language instruction $l$ to a chunk of $T$ consecutive actions $\mathbf{a}_{t:t+T} \in \mathbb{R}^{T \times D}$, where $D$ is the action dimensionality. Following the action-chunking paradigm~\cite{black2024pi0,lightvla}, predicting the full chunk jointly at each control step enables temporally coherent execution over the horizon. A pretrained VLM serves as the backbone:  the vision encoder tokenizes $o_t$ into $M$ spatial image tokens of hidden dimension $d$; together with the tokenized instruction $l$, they are fed into the VLM, which produces contextual representations $\mathbf{c}$ that condition the action head.                                                                         
\paragraph{Flow Matching Action Head.}
The action head is trained via conditional flow matching~\cite{black2024pi0, nvidia2025groot}. A linear interpolation path is defined between standard Gaussian noise $\mathbf{x}_0 \sim \mathcal{N}(\mathbf{0}, \mathbf{I})$ and the ground-truth action chunk $\mathbf{x}_1 = \mathbf{a}_{t:t+T}$:
\begin{equation}
    \mathbf{x}_\tau = (1-\tau)\,\mathbf{x}_0 + \tau\,\mathbf{x}_1, \quad \tau \sim \mathcal{U}[0,1].
    \label{eq:fm_path}
\end{equation}
A DiT network $v_\theta(\mathbf{x}_\tau, \tau, \mathbf{c})$ is trained to match the constant conditional velocity $\mathbf{x}_1 - \mathbf{x}_0$:
\begin{equation}
    \mathcal{L}_{\text{action}} = \mathbb{E}_{\tau,\,\mathbf{x}_0,\,\mathbf{x}_1}\!\left[\left\|v_\theta(\mathbf{x}_\tau,\tau,\mathbf{c}) - (\mathbf{x}_1 - \mathbf{x}_0)\right\|^2\right].
    \label{eq:fm_loss}
\end{equation}
At inference, the action chunk is recovered by integrating $\mathrm{d}\mathbf{x}/\mathrm{d}\tau = v_\theta(\mathbf{x}_\tau, \tau, \mathbf{c})$ from $\tau{=}0$ to $\tau{=}1$.

\subsection{Point Tracking}
\label{sec:cotracker}

To ground the VLM's image token representations in explicit spatial motion, we supervise them to predict 2D point trajectories spanning the entire action chunk. A key prerequisite is that these representations be \emph{goal-aware}: the model must understand where objects should move given the language instruction. In standard autoregressive VLMs with causal attention, image tokens placed before text tokens cannot attend to text tokens that appear later in the sequence. We therefore reorder the input sequence such that text tokens precede image tokens, thereby ensuring that the hidden state of each image token is conditioned on the complete language instruction, which provides a necessary foundation for decoding task-relevant motion from image-token hidden states.

\paragraph{Sparse Point Selection.}
The vision encoder processes the input image at a fixed patch stride $s$, producing an $H' \times W'$ spatial feature map. We define a sparse set of $N = H' \times W'$ query points by mapping the center of each compressed patch cell back to its corresponding coordinate in the original image, establishing a one-to-one correspondence between image tokens and spatial query locations, yielding a lightweight geometric supervision signal. Each image token's hidden state is thus directly associated with a trackable point over the action chunk horizon.

\paragraph{Supervision via Point Tracking.}
Given the current frame $o_t$ and the $N$ grid points, a frozen point tracker serves as a teacher to generate sparse ground-truth displacement targets, , avoiding the need for costly dense pixel rendering. $\mathbf{d}^{\star} \in \mathbb{R}^{N \times T \times 2}$ (normalized by image dimensions) and visibility labels $\mathbf{v}^{\star} \in \{0,1\}^{N \times T}$, where $T$ is the action chunk length. Concretely, we use CoTracker-v3~\cite{karaev2024cotracker} as the tracking teacher. The tracker follows each point forward across all $T$ frames of the action chunk, providing dense per-frame temporal supervision. It is used exclusively during training and discarded at inference; all motion knowledge is absorbed into the VLM backbone parameters.

\paragraph{Alignment Heads.}
From the VLM hidden states at the $N$ image token positions, we extract latent features $\mathbf{h} \in \mathbb{R}^{N \times d}$. Two lightweight projection heads predict point displacements and visibility:
\begin{equation}
    \hat{\mathbf{d}} = f_{\text{track}}(\mathbf{h}) \in \mathbb{R}^{N \times T \times 2}, \quad
    \hat{\mathbf{v}} = f_{\text{vis}}(\mathbf{h}) \in \mathbb{R}^{N \times T}
    \label{eq:track_heads}
\end{equation}
where $f_{\text{track}}$ is a two-layer MLP (LayerNorm $\to$ Linear $\to$ GELU $\to$ Linear) and $f_{\text{vis}}$ has a similar architecture with smaller hidden dimension.

\paragraph{Motion Losses.}
The motion alignment objective combines displacement regression, visibility classification, and trajectory smoothness:
\begin{equation}
    \mathcal{L}_{\text{track}}
    =
    \mathcal{L}_{\text{disp}}
    +
    \mathcal{L}_{\text{vis}}
    +
    \lambda_{\text{smooth}}\mathcal{L}_{\text{smooth}}
\end{equation}
\begin{align}
    \mathcal{L}_{\text{disp}}
    &=
    \frac{1}{NT}
    \sum_{n,t}
    v^{\star}_{n,t}
    \|
    \hat{\mathbf{d}}_{n,t}-\mathbf{d}^{\star}_{n,t}
    \|^2 \\
    \mathcal{L}_{\text{vis}}
    &=
    \mathrm{BCE}
    (
    \hat{\mathbf{v}},
    \mathbf{v}^\star
    ) \\
    \mathcal{L}_{\text{smooth}}
    &=
    \frac{1}{N(T-2)}
    \sum_{n,t}
    v^{\star}_{n,t}
    \|
    \Delta^2 \hat{\mathbf{d}}_{n,t}
    \|^2
\end{align}
The smoothness term regularizes the second-order temporal difference $\Delta^2 \hat{\mathbf{d}}_{n,t} = \hat{\mathbf{d}}_{n,t+1} - 2\hat{\mathbf{d}}_{n,t} + \hat{\mathbf{d}}_{n,t-1}$, suppressing abrupt motion changes and promoting physically plausible trajectories. All losses are masked by ground-truth visibility so that occluded points do not contribute noisy gradients.

\subsection{Future Feature Prediction}
\label{sec:foresight}

While the tracking branch provides sparse per-frame motion supervision grounded in individual spatial locations, it does not encode a holistic representation of the final goal state. Therefore, we append $K$ special \texttt{<Foresight>} tokens after the image and instruction tokens, and train their hidden states to encode the visual features of $o_{t+T}$, the final frame of the action chunk. 
These tokens thus provide a compact semantic representation of the target scene configuration. 
Unlike the tracking branch which uses all $N$ image token hidden states to predict per-frame displacements across $T$ steps, the $K$-token bottleneck is deliberately compact, encouraging the model to capture global scene-level changes rather than fine-grained temporal details.

\paragraph{EMA Teacher.}
To provide stable regression targets, we maintain a shadow copy of the vision encoder updated via exponential moving average (EMA) with momentum $\mu$ as the teacher. For the target frame $o_{t+T}$, the frozen EMA teacher yields target visual features $\mathbf{z}^{\star} \in \mathbb{R}^{M \times d}$, where $M = H' \times W'$ is the number of visual patches.

\paragraph{MAE Decoder.}
To reconstruct the compact $M$-patch target from the compact $K$-token bottleneck, we employ a lightweight MAE decoder. The $K$ \texttt{<Foresight>} hidden states are concatenated with $M{-}K$ learnable mask tokens and passed through a lightweight ViT decoder with sinusoidal positional embeddings, yielding reconstructed predictions $\hat{\mathbf{z}} \in \mathbb{R}^{M \times d}$. The reconstruction objective is a cosine-similarity loss:
\begin{equation}
    \mathcal{L}_{\text{foresight}} = 1 - \frac{1}{M}\sum_{i=1}^{M} \frac{\hat{\mathbf{z}}_i \cdot \mathbf{z}^{\star}_i}{\|\hat{\mathbf{z}}_i\| \, \|\mathbf{z}^{\star}_i\|}.
    \label{eq:foresight_loss}
\end{equation}

For inference, the $K$ \texttt{<Foresight>} tokens are retained in the input sequence, allowing the action head to cross-attend to future-informed hidden states and acquire an explicit prior of the target visual state for the subsequent action chunk.

\subsection{Future-Conditioned Point Tracking}
\label{sec:goal_cross_attn}

The two aforementioned objectives, namely future feature prediction and geometric point tracking, can be optimized independently as parallel auxiliary loss terms. However, this treats them as unrelated regularizers. We hypothesize that motion prediction benefits from knowing the \emph{intended} future state: if the model knows \emph{what} the future visual state will look like, predicting \emph{how} each point moves to reach it becomes a better-constrained problem.

To instantiate this conditioning, we insert a lightweight FCCA module between the VLM's visual token extraction and the displacement projector. Let $\mathbf{H}_{\text{vis}} \in \mathbb{R}^{N \times d}$ denote the vision features at grid positions and $\mathbf{H}_{\text{fut}} \in \mathbb{R}^{K \times d}$ the VLM hidden states at \texttt{<Foresight>} positions. The FCCA module computes:
\begin{equation}
    \tilde{\mathbf{H}}_{\text{vis}} = \mathbf{H}_{\text{vis}} + \operatorname{MHA}(\operatorname{LN}(\mathbf{H}_{\text{vis}}),\,\operatorname{LN}(\mathbf{H}_{\text{fut}}),\,\operatorname{LN}(\mathbf{H}_{\text{fut}}))
    \label{eq:fcca}
\end{equation}
where MHA is multi-head attention with 8 heads and $\text{LN}$ denotes layer normalization.

The output projection of the MHA is \emph{zero-initialized}, such that the module behaves as an identity mapping at the beginning of training and thereby preserves the pretrained VLM features without perturbation. As training progresses, the module gradually learns to inject future information into the spatial tokens that feed the displacement predictor.

\begin{table*}[!t]
\centering
\small
\setlength{\tabcolsep}{12pt}
\begin{tabular}{clccccr}
\toprule
\textbf{Type} & \textbf{Method} & \textbf{Spatial} & \textbf{Object} & \textbf{Goal} & \textbf{Long} & \textbf{Avg.} \\
\midrule
\multirow{8}{*}{VLA/WAM}
& GR00T-N1.5~\cite{nvidia2025groot} \color{gray}(arXiv '25)  & 92.0 & 92.0 & 86.0 & 76.0 & 86.5 \\
& $\pi_0$~\cite{black2024pi0} \color{gray}(RSS '25)          & 96.8 & 98.8 & 95.8 & 85.2 & 94.1 \\
& X-VLA~\cite{zheng2025xvla} \color{gray}(ICLR '26) &98.2 &98.6 &97.8 &97.6 & 98.1 \\
& LangForce~\cite{LangForce_2026_ICML} \color{gray}(ICML '26) &99.2 &99.6 &99.4 &95.2 &98.4 \\
& Fast-WAM~\cite{yuan2026fastwam} \color{gray}(arXiv '26) &98.2 &\textbf{100.0} &97.0 &95.2 &97.6 \\
& LingBot-VA~\cite{li2026causalworldmodelingrobot} \color{gray}(RSS '26) &98.5 &99.6 &97.2 &98.5 &98.5 \\
& Cosmos Policy~\cite{kim2026cosmospolicy} \color{gray}(ICLR '26) &98.1 &\textbf{100.0} &98.2 &\textbf{97.6} &98.5 \\
& Spatial Forcing~\cite{li2026spatialforcing} \color{gray}(ICLR '26)  & \textbf{99.4} & 99.6 & 98.8 & 96.0 & 98.5 \\

\midrule
\multirow{5}{*}{Future Prediction}
& WorldVLA~\cite{cen2025worldvla} \color{gray}(arXiv '25)       & 87.6 & 96.2 & 83.4 & 60.0 & 81.8 \\
& DreamVLA~\cite{zhang2025dreamvla} \color{gray}(NeurIPS '25)   & 97.5 & 94.0 & 89.5 & 89.5 & 92.6 \\
& UniVLA~\cite{bu2025univla} \color{gray}(RSS '25)          & 96.5 & 96.8 & 95.6 & 92.0 & 95.2 \\
& LaRA-VLA~\cite{lara2026} \color{gray}(ICML '26)            & 96.4 & 99.8 & 98.6 & 96.6 & 97.9 \\
& HiF-VLA~\cite{HiF-VLA} \color{gray}(CVPR '26)            & 98.8 & 99.4 & 97.4 & 96.4 & 98.0 \\
\midrule
\multirow{3}{*}{Point Tracking}
& FlowVLA~\cite{zhong2025flowvla} \color{gray}(arXiv '25)          & 93.2 & 95.0 & 91.6 & 72.6 & 88.1 \\
& GeoPredict~\cite{qian2026geopredict} \color{gray}(CVPR '26)      & 98.0 & 98.2 & 95.7 & 94.0 & 96.5 \\
& JOPAT~\cite{guan2026jopat} \color{gray}(arXiv '26)                   & 97.2 & 98.9 & 98.4 & 96.4 & 97.8 \\

\midrule
\multirow{5}{*}{Ours}
& Base Backbone                                                   & 97.8 & 98.8 & 97.4 & 92.0 & 96.5 \\
& + Future Prediction                                            & 99.0 & 99.4 & 97.2 & 94.4 & 97.5 \\
& + Tracking                                          & 98.6 & 99.2 & 99.0 & 94.4 & 97.8 \\
& + Future Prediction + Tracking 
& 98.8 & 99.4 & 99.2 & 95.8 & 98.3 \\
& + Future Prediction + Tracking + FCCA                         & 98.4 & 99.6 & \textbf{99.4} & \textbf{97.6} & \textbf{98.8} \\
\bottomrule
\end{tabular}
\caption{LIBERO evaluation results. We compare our method with representative VLA baselines, future-prediction methods, and point tracking methods, and further provide ablations of each proposed component. The full model achieves the highest average success rate across 4 task suites.}
\label{tab:main}
\end{table*}

\begin{table*}[ht]
\centering
\setlength{\tabcolsep}{1.5pt}
\begin{tabular}{lccccccccc}
\toprule
\textbf{Method} & \textbf{Pretrain} &
\textbf{Camera} & \textbf{Robot} & \textbf{Language} &
\textbf{Light} & \textbf{Background} &
\textbf{Noise} & \textbf{Layout} & \textbf{Total} \\
\midrule
$\pi_0$~\cite{black2024pi0}
& \textcolor{poscolor}{\ding{51}}
& 13.8 & 6.0  & 58.8 & 85.0 & 81.4 & 79.0 & 68.9 & 53.6 \\

VLA-JEPA~\cite{yang2026abotm0}
& \textcolor{poscolor}{\ding{51}}
& 63.3 & 67.1 & 85.4 & 95.6 & 93.6 & 66.3 & 85.1 & 79.5 \\

Abot-M0~\cite{yang2026abotm0}
& \textcolor{poscolor}{\ding{51}}
& 60.4 & 67.9 & 86.4 & 96.2 & 91.6 & 86.4 & 82.6 & 80.5  \\ \midrule

OpenVLA~\cite{kim2024openvla}
& \textcolor{negcolor}{\ding{55}}
& 0.8 & 3.5 & 23.0 & 8.1 & 34.8 & 15.2 & 28.5 & 15.6 \\

WorldVLA~\cite{cen2025worldvla}
& \textcolor{negcolor}{\ding{55}}
& 0.1 & 27.9 & 41.6 & 43.7 & 17.1 & 10.9 & 38.0 & 25.0 \\

UniVLA~\cite{bu2025univla}
& \textcolor{negcolor}{\ding{55}}
& 1.8 & 46.2 & 69.6 & 69.0 & 81.0 & 21.2 & 31.9 & 42.9 \\

DreamVLA~\cite{zhang2025dreamvla}
& \textcolor{negcolor}{\ding{55}}
& 26.2 & 17.6 & 67.0 & 77.5 & 71.5 & 53.6 & 43.5 & 48.9 \\

Cosmos Policy~\cite{kim2026cosmospolicy}
& \textcolor{negcolor}{\ding{55}}
& 69.6 & 51.0 & 89.6 & 97.7 & 85.7 & 87.3 & 83.7 & 79.7 \\

StarVLA~\cite{ye2026starvla}
& \textcolor{negcolor}{\ding{55}}
& 52.5 & 49.8 & 88.5 & 95.7 & 95.7 & 73.0 & 76.9 & 74.1 \\

Qwen-RobotManip-scratch~\cite{yuan2026qwenrobotmanip}
& \textcolor{negcolor}{\ding{55}}
& \textbf{70.4} & 44.9 & 88.1 & \textbf{95.8} & 95.5 & \textbf{84.4} & 79.1 & 78.3 \\
\midrule
FoMoVLA (Ours)
& \textcolor{negcolor}{\ding{55}}
& 64.0 & \textbf{62.2} & \textbf{94.0} & 94.1 & \textbf{96.2} & 82.2 & \textbf{79.6} & \textbf{80.5} \\
\bottomrule
\end{tabular}%
\caption{
Out-of-distribution robustness evaluation on LIBERO-Plus. All methods are evaluated zero-shot without fine-tuning.
}
\label{tab:liberoplus}
\end{table*}

This design connects the two branches by using the foresight representation as a conditioning signal to motion prediction via FCCA, ensuring that the predicted future state and point trajectories are jointly consistent.

\subsection{Training and Inference}
\label{sec:training}

\paragraph{Training Objective.}
The overall training loss combines action prediction with the two auxiliary objectives:
{\small
\begin{equation}
    \mathcal{L} = \mathcal{L}_{\text{action}} + \lambda_1 \mathcal{L}_{\text{foresight}} + \lambda_2 \left(\mathcal{L}_{\text{disp}} + \mathcal{L}_{\text{vis}} + \lambda_{\text{smooth}} \mathcal{L}_{\text{smooth}}\right)
    \label{eq:total_loss}
\end{equation}
}
where $\lambda_1$, $\lambda_2$, and $\lambda_{\text{smooth}}$ are loss weighting coefficients detailed in Section~\ref{sec:impl}. The action loss $\mathcal{L}_{\text{action}}$ is the flow-matching objective of the DiT action head. The point tracker teacher is used only to generate supervisory signals and receives no gradient updates; the EMA teacher is updated solely via momentum averaging without backpropagation.

\paragraph{Inference.}
At test time, all auxiliary branches are discarded. The only modification is the addition of $K$ \texttt{<Foresight>} tokens to the input, increasing median latency by 9.4 ms and GPU memory by 0.1 GB (Table~\ref{tab:cost}).

\section{Experiments}

\subsection{Experimental Setup}

\paragraph{Benchmarks.}
\textbf{LIBERO}~\cite{liu2023libero} comprises four suites (Spatial, Object, Goal, Long), each with 10 tasks and 50 demonstrations per task. We report success rates over 20 rollouts per task with $T{=}8$.
\textbf{RoboCasa GR-1 Tabletop}~\cite{nasiriany2024robocasa} contains 24 tabletop pick-place tasks for the GR-1 humanoid, with 1,000 demonstrations per task (24,000 total). We report success rates over 50 rollouts per task with $T{=}16$.
\textbf{LIBERO-Plus}~\cite{fei2025liberoplus} extends LIBERO with 7 perturbation dimensions (viewpoint, robot state, language, lighting, background, noise, layout) across 10,030 instances. We evaluate zero-shot without any fine-tuning on the perturbed data.

\paragraph{Baselines.} We compare FoMoVLA with three groups of existing methods. 
The first group consists of generalist VLA and world-action models (GR00T-N1.5~\cite{nvidia2025groot} and $\pi_0$~\cite{black2024pi0}, X-VLA~\cite{zheng2025xvla}, LangForce~\cite{LangForce_2026_ICML}, Spatial Forcing~\cite{li2026spatialforcing}, Fast-WAM~\cite{yuan2026fastwam}, LingBot-VA~\cite{li2026causalworldmodelingrobot}, Cosmos Policy~\cite{kim2026cosmospolicy}). The second group includes future prediction methods (WorldVLA~\cite{cen2025worldvla}, DreamVLA~\cite{zhang2025dreamvla}, UniVLA~\cite{bu2025univla}, and LaRA-VLA~\cite{lara2026}, HiF-VLA~\cite{HiF-VLA}) that predict future visual representations without modeling geometric motion. The third group contains point tracking methods (FlowVLA~\cite{zhong2025flowvla}, GeoPredict~\cite{qian2026geopredict} and JOPAT~\cite{guan2026jopat}), which leverage optical flow or point trajectories but do not predict future visual states.

\paragraph{Implementation Details.}\label{sec:impl}
All variants are built upon the StarVLA-GR00T~\cite{ye2026starvla} as base backbone. 
Input images are resized to $224{\times}224$ pixels. The ViT patch size is 16, yielding a $14{\times}14$ feature map that is further compressed by the spatial merger module to $8{\times}8$ image tokens ($M{=}N{=}64$) fed to the VLM. 
The $N{=}64$ tracking query points are placed at the center of each $8{\times}8$ patch cell mapped back to the original image resolution. 
We use $K{=}16$ \texttt{<Foresight>} tokens. 
The MAE decoder expands these with $M{-}K{=}48$ learnable mask tokens through a 2-layer ViT decoder. 
The EMA teacher uses momentum $\mu{=}0.999$. The loss weights are $\lambda_1{=}0.1$, $\lambda_2{=}0.3$, and $\lambda_{\text{smooth}}{=}0.1$. 
We train with 8$\times$H20 GPUs with DeepSpeed ZeRO-2, AdamW ($\beta_1{=}0.9$, $\beta_2{=}0.95$), and per-module learning rates (VLM: $1{\times}10^{-5}$; auxiliary heads: $1{\times}10^{-4}$). 
For LIBERO, the training steps are 30K  with batch size fixed to 12 while for RoboCasa, the training steps are 100K with the batch size fixed to 8.

\subsection{Main Experiments}

Tab.~\ref{tab:main} presents the main results and several observations are as follows:
(1)~Both auxiliary objectives individually improve over the Base VLA, confirming that predictive and geometric supervision provide complementary inductive biases.
(2)~Naively combining both objectives without future interaction yields modest improvements, suggesting that independent multi-task training suffers from limited synergy.
(3)~The full design with FCCA achieves the competitive results across all LIBERO suites, demonstrating that explicitly connecting the two objectives via future conditioning produces gains beyond simple loss aggregation.
The advantage is evident on LIBERO-Long and RoboCasa, where long-horizon reasoning and precise spatial control are more critical.

Tab.~\ref{tab:liberoplus} reports zero-shot OOD robustness on LIBERO-Plus. FoMoVLA achieves 80.5\% overall, matching Abot-M0 without additional pretrain and significantly outperforming StarVLA (+6.4\%). The strongest gains appear on language (+5.5\% over StarVLA) and background perturbations. The gains on camera and robot state shifts are more modest, likely because the model is trained on fixed viewpoints and initial poses, making it difficult to generalize to novel configurations with purely 2D motion cues.

Tab.~\ref{tab:robocasa} summarizes average success rates on RoboCasa GR-1 Tabletop. FoMoVLA achieves 56.9\%, improving over the StarVLA-GR00T backbone by 9.1\%, demonstrating consistent gains from our dual-objective training on a humanoid platform with egocentric observations. Per-task results and component-wise ablation on RoboCasa are provided in Tab.~\ref{tab:robocasa_ablation} in the appendix.
\begin{table}[ht]
  \centering
  \small
  \setlength{\tabcolsep}{20pt}
  \begin{tabular}{lc}
  \toprule
  \textbf{Method} & \textbf{Avg.} \\
  \midrule
  $\pi_{0.5}$~\cite{black2025pi05}      & 37.0 \\
  StarVLA-FAST~\cite{ye2026starvla}      & 39.0 \\
  StarVLA-$\pi$~\cite{ye2026starvla}    & 43.9 \\
  StarVLA-GR00T~\cite{ye2026starvla}    & 47.8 \\
  GR00T-N1.6~\cite{nvidia2025groot}     & 47.6 \\
  StarVLA-OFT~\cite{ye2026starvla}      & 48.8 \\
  \midrule
  FoMoVLA (Ours)                       & \textbf{56.9} \\
  \bottomrule
  \end{tabular}
  \caption{Average success rate (\%) on RoboCasa GR-1 Tabletop
           (24 tasks, 50 rollouts/task).}
  \label{tab:robocasa}
\end{table}

\subsection{Ablation Study}

\begin{figure}[ht]
    \centering
    \includegraphics[width=\columnwidth]{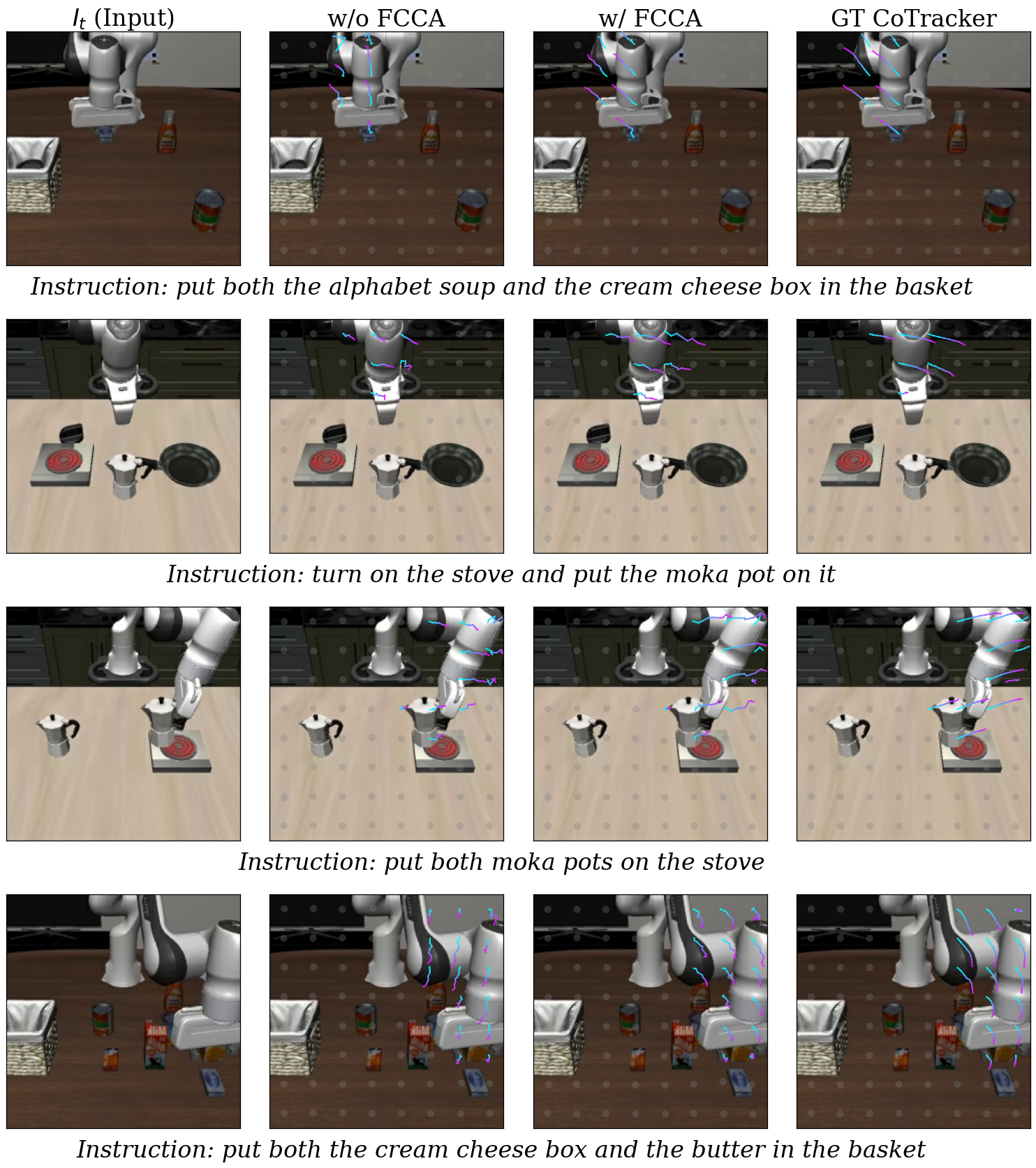}
    \caption{Effect of FCCA on point tracking. Conditioning point tracking on visual foresight yields trajectories aligned with action dynamics.}
    \label{fig:fcca_cotracker}
\end{figure}

\paragraph{Future Conditioning.}
Tab.~\ref{tab:main} shows that the full model with FCCA achieves consistent gains over the non-FCCA variant (e.g., +1.8\% in LIBERO-Long), confirming that the two objectives achieve synergy only when motion prediction is explicitly conditioned on the predicted future representation. Fig.~\ref{fig:fcca_cotracker} illustrates this qualitatively: without FCCA, predicted point trajectories exhibit incorrect displacement directions; with FCCA, motion predictions align with the actual tracking trajectories of the points.This qualitative evidence suggests that jointly optimizing the two objectives leads to more coherent spatiotemporal representations than naive multi-task training.

\begin{figure}[!t]
    \centering
    \includegraphics[width=\columnwidth]{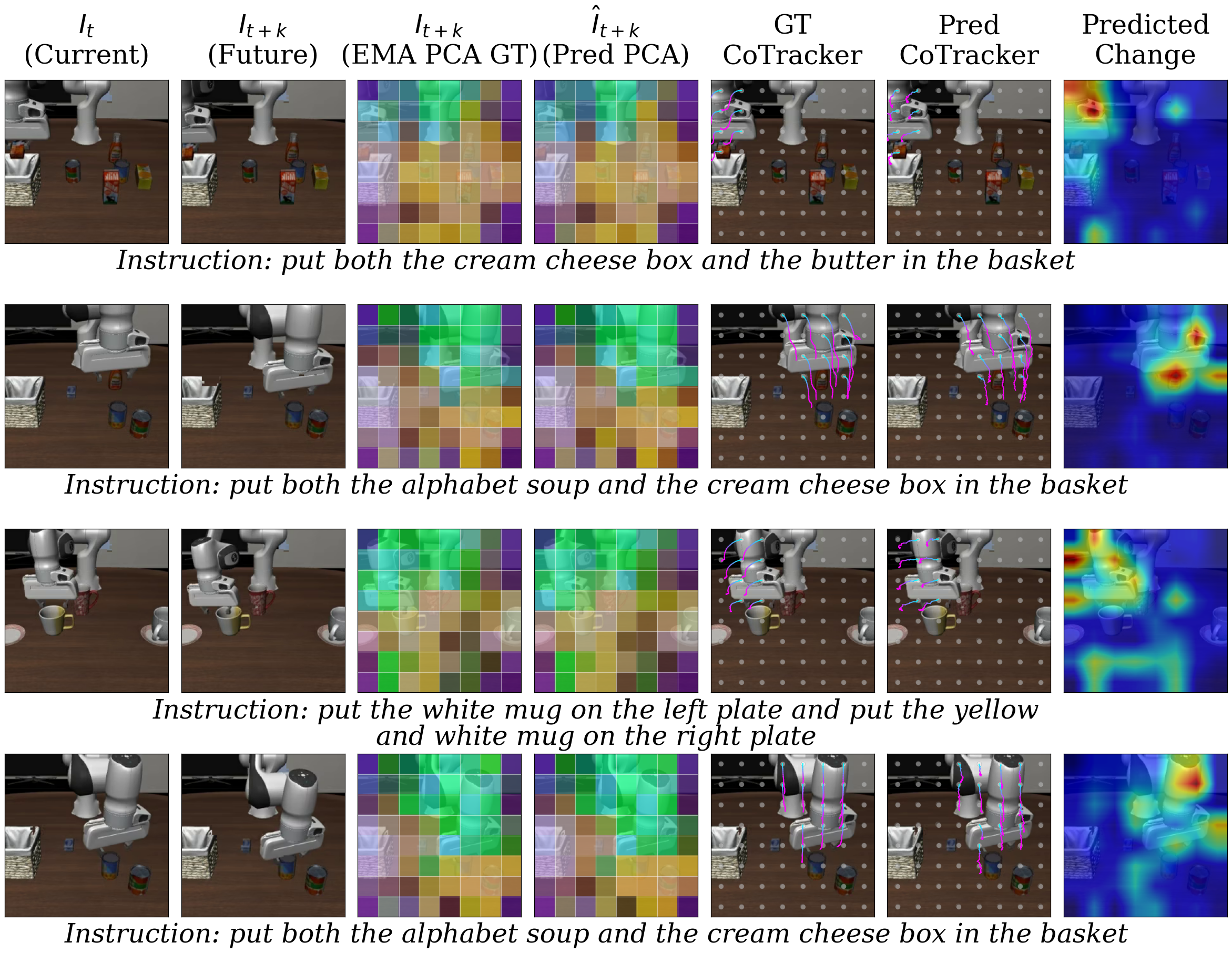}
    \caption{Qualitative results of FoMoVLA. 
FoMoVLA effectively learns future features and point tracking.
    }
    \label{fig:qualitative}
\end{figure}

\paragraph{Sparse Supervision Design.}

We first ablate the point grid density (top section of Table~\ref{tab:design}):
the sparse $8\times8$ grid (64 points, patch-aligned) achieves the best overall
performance, particularly on long-horizon tasks, while denser grids ($16\times16$) degrade performance. This validates our design choice of lightweight sparse point supervision, which uses a $4\times$ sparser set yet sufficiently captures task-relevant motion.

For the foresight decoder architecture, the MAE bottleneck design outperforms
Direct-64 on long-horizon tasks, confirming that the compact $K$-token bottleneck
encourages global goal-state encoding rather than patch-level memorization.

\paragraph{Auxiliary Task Coupling.}
  \begin{table}[ht]                                                                                            
  \centering      
  \small                                                                                                                            
  \setlength{\tabcolsep}{3.5pt}                                                                                                      
\begin{tabular}{lccccc}
  \toprule
  \textbf{Design} & \textbf{Spatial} & \textbf{Object} & \textbf{Goal} & \textbf{Long} & \textbf{Avg.} \\
  \midrule
  \multicolumn{6}{>{\columncolor{gray!30}}l}{\emph{Point Grid Density}} \\
  $8{\times}8$ (64 points) & 98.4 & \textbf{99.6} & 99.4 & \textbf{97.6}  & \textbf{98.8} \\
  $16{\times}16$ (256 points) & \textbf{98.8} & 99.4 & \textbf{99.6} & 94.8  & 98.1 \\
  \midrule
 \multicolumn{6}{>{\columncolor{gray!30}}l}{\emph{Foresight Architecture}}\\
  MAE (16$\to$64 tokens) & 98.4 & \textbf{99.6} & \textbf{99.4} & \textbf{97.6} & \textbf{98.8} \\
  Direct-64 (64 tokens) & \textbf{98.6} & 99.4 & 98.0 & 95.0 & 97.8  \\
  \midrule
  \multicolumn{6}{>{\columncolor{gray!30}}l}{\emph{Foresight–Motion Coupling}} \\
  Shared goal tokens          & 98.2 & 99.2 & 96.0 & 85.8 & 94.8 \\
  Separate query tokens       & 98.4 & \textbf{99.6} & 99.0 & 95.8 & 98.2 \\
  Shared image tokens         & 97.8 & 99.0 & 97.4 & 95.6 & 97.4 \\
  Image + Goal CrossAttn      & \textbf{98.4} & \textbf{99.6} & \textbf{99.4} & \textbf{97.6} & \textbf{98.8} \\
  \bottomrule
  \end{tabular}
  \caption{Design choices: Point Grid Density, Foresight Architecture and Foresight-Motion Coupling design on LIBERO success rate (\%).}
  \label{tab:design}
\end{table}

The bottom section of Table~\ref{tab:design} ablates design choices for coupling
the foresight and tracking objectives. When both branches draw from the same 16 learnable goal tokens (\emph{Shared goal tokens}), performance degrades drastically on long-horizon tasks (85.8\%). We attribute this drop to gradient interference on the shared token embeddings, as the two auxiliary objectives favor different information: future appearance versus spatial motion. Assigning each branch its own set of 16 learnable queries (\emph{Separate query tokens}) eliminates this gradient conflict and recovers
strong performance (98.2\%). Using the 64 ViT image tokens for both branches (\emph{Shared image tokens}) offers a different solution: since these features are primarily optimized by the main by the main action prediction objective, the auxiliary losses no longer directly clash at
the embedding level, yielding 97.4\%. Our final design (\emph{Image + Goal CrossAttn}) achieves the best result (98.8\%) by keeping image tokens as the motion decoder input while injecting goal-oriented context via cross-attention from the foresight query tokens, which enables cooperative information flow without representational interference.

\subsection{Scalability Across Different Frameworks.}
  To evaluate whether FoMoVLA is tied to a specific action formulation, we further instantiate it on top of two alternative StarVLA variants: StarVLA-$\pi$ and StarVLA-OFT. As
  shown in Tab.~\ref{tab:policy_heads}, FoMoVLA consistently improves both variants. In particular, it boosts StarVLA-$\pi$ from 95.7\% to 97.9\% average success rate, with the largest gain on LIBERO-Long (+8.2), while improving StarVLA-OFT from 96.6\% to 98.0\%. These results indicate that FoMoVLA complements different policy heads rather than relying on a specific action decoder design.

\begin{table}[t]
  \centering
  \small
  \setlength{\tabcolsep}{4pt}
  \begin{tabular}{lccccc}
  \toprule
  \textbf{Method} & \textbf{Spatial} & \textbf{Object} & \textbf{Goal} & \textbf{Long} & \textbf{Avg.} \\
  \midrule
  StarVLA-$\pi$ & 98.8 & 99.6 & 95.8 & 88.4 & 95.7 \\
  + FoMoVLA & 98.2 & 98.6 & 98.0 & 96.6 & 97.9 \\
  $\Delta$ & \textbf{\textcolor{negcolor}{-0.6}} & \textbf{\textcolor{negcolor}{-1.0}} & \textbf{\textcolor{poscolor}{+2.2}} & \textbf{\textcolor{poscolor}{+8.2}} & \textbf{\textcolor{poscolor}{+2.2}} \\
  \midrule
  StarVLA-OFT & 97.8 & 98.6 & 96.2 & 93.8 & 96.6 \\
  + FoMoVLA & 99.0 & 99.4 & 99.0 & 94.8 & 98.0 \\
  $\Delta$ & \textbf{\textcolor{poscolor}{+1.2}} & \textbf{\textcolor{poscolor}{+0.8}} & \textbf{\textcolor{poscolor}{+2.8}} & \textbf{\textcolor{poscolor}{+1.0}} & \textbf{\textcolor{poscolor}{+1.4}} \\
  \bottomrule
  \end{tabular}
  \caption{Scalability of FoMoVLA across diverse policy heads of mainstream VLA frameworks.}
  \label{tab:policy_heads}
\end{table}

\subsection{Qualitative Analysis.}
We visualize current observation features, predicted vs.\ GT future features, predicted vs.\ GT point tracking, and the predicted change heatmap (per-patch cosine distance between current and predicted future features) in Fig.~\ref{fig:qualitative}. The predicted foresight closely reconstructs GT future features, confirming that the $K$-token bottleneck captures global scene-level changes rather than memorizing local texture. The predicted point trajectories align with GT motion, showing that geometric supervision is successfully absorbed into the VLM backbone. Critically, the predicted change heatmap concentrates on task-relevant regions (manipulated object and end-effector path) while suppressing static background. The foresight anchor tells the model \emph{where} the scene changes and point tracking guide \emph{how}, together providing complementary evidence that FoMoVLA learns a spatio-temporal grounded representation rather than independently optimizing two auxiliary losses.

\section{Conclusion}
In this paper, we proposed FoMoVLA, a training-only framework that equips VLA models with explicit spatio-temporal foresight through future feature prediction and future-conditioned point tracking. jointly learning compact future feature prediction and sparse, lightweight point tracking, our method produces more predictive and spatio-temporally grounded visual representations, leading to improved modeling of future state transitions and more effective robotic manipulation. Since all auxiliary components are used only during training, FoMoVLA incurs no additional inference overhead. Extensive experiments on LIBERO and RoboCasa demonstrate consistent improvements over strong baselines, highlighting the effectiveness of explicit spatio-temporal supervision for VLA models.

Regarding limitations, current tracking supervision only models view dependent motion in image space, failing to capture the 3D geometry over a dynamic scene.
For future works, a promising direction is to complement 2D point tracking with 3D motion prediction, enabling richer geometric reasoning and greater robustness to viewpoint and depth variation. 

\bibliography{references}
\clearpage
\twocolumn[
\begin{center}
{\Large\textbf{FoMoVLA: Bridging Visual Foresight and Motion Guidance for Vision-Language-Action Models} \\[0.5em]
\large Supplementary Material}
\end{center}
\vspace{1em}
]
\appendix
\section{}
\subsection{Model Parameter Breakdown}

Tab.~\ref{tab:params} details the parameter breakdown of FoMoVLA. During training, the full model introduces 60.7\,M additional parameters (+1.3\%) from auxiliary branches: foresight decoder (30.5\,M), FCCA module (26.2\,M), track projector (2.6\,M), and visibility head (1.3\,M). Since all auxiliary branches are discarded at inference, the deployment parameters remain identical to the Vanilla baseline (4599.3\,M).

\begin{table}[ht]
\centering
\small
\begin{tabular}{lcc}
\toprule
\textbf{Module} & \textbf{Vanilla} & \textbf{FoMoVLA} \\
\midrule
VLM Backbone         & 4437.8 & 4437.8 \\
Action Model         & 161.5  & 161.5 \\
Foresight Decoder    & --     & 30.5 \\
FCCA Module          & --     & 26.2 \\
Track Projector      & --     & 2.6 \\
Visibility Head      & --     & 1.3 \\
\midrule
Training Total (M)   & 4599.3 & 4660.0 \\
Inference Total (M)  & 4599.3 & 4599.3 \\
\bottomrule
\end{tabular}
\caption{Model parameter breakdown (in millions). Auxiliary branches (shaded rows) are discarded at inference, keeping the deployment footprint identical to the Vanilla baseline.}
\label{tab:params}
\end{table}

\subsection{Attention Mask Design}
\label{sec:appendix_attention}

As described in Section~\ref{sec:cotracker}, we reorder the input sequence to place instruction tokens before image tokens to ensure goal-aware visual representations. Here we detail the full attention mask design governing inter-modal interactions within the VLM backbone.

The input sequence is organized as: \textbf{Instruction} $\to$ \textbf{Current Image} $\to$ \textbf{Foresight} $\to$ \textbf{Action}. We employ a structured attention mask with the following rules: (1)~Instruction tokens use strict causal attention among themselves, attending only to preceding tokens; (2)~Current image tokens attend causally to all preceding instruction tokens and among themselves, ensuring that visual representations are fully conditioned on the language instruction; (3)~Foresight tokens employ bidirectional attention among themselves, while additionally attending to all instruction and current image tokens---this allows the $K$ \texttt{<Foresight>} tokens to freely exchange information, facilitating holistic encoding of the predicted future state grounded in both the task specification and the current observation; and (4)~Action tokens attend to all preceding modalities---instruction, current image, and foresight tokens---enabling the action head to leverage the complete context including future-informed representations for action prediction.

\begin{figure}[t]
    \centering
    \includegraphics[width=\columnwidth]{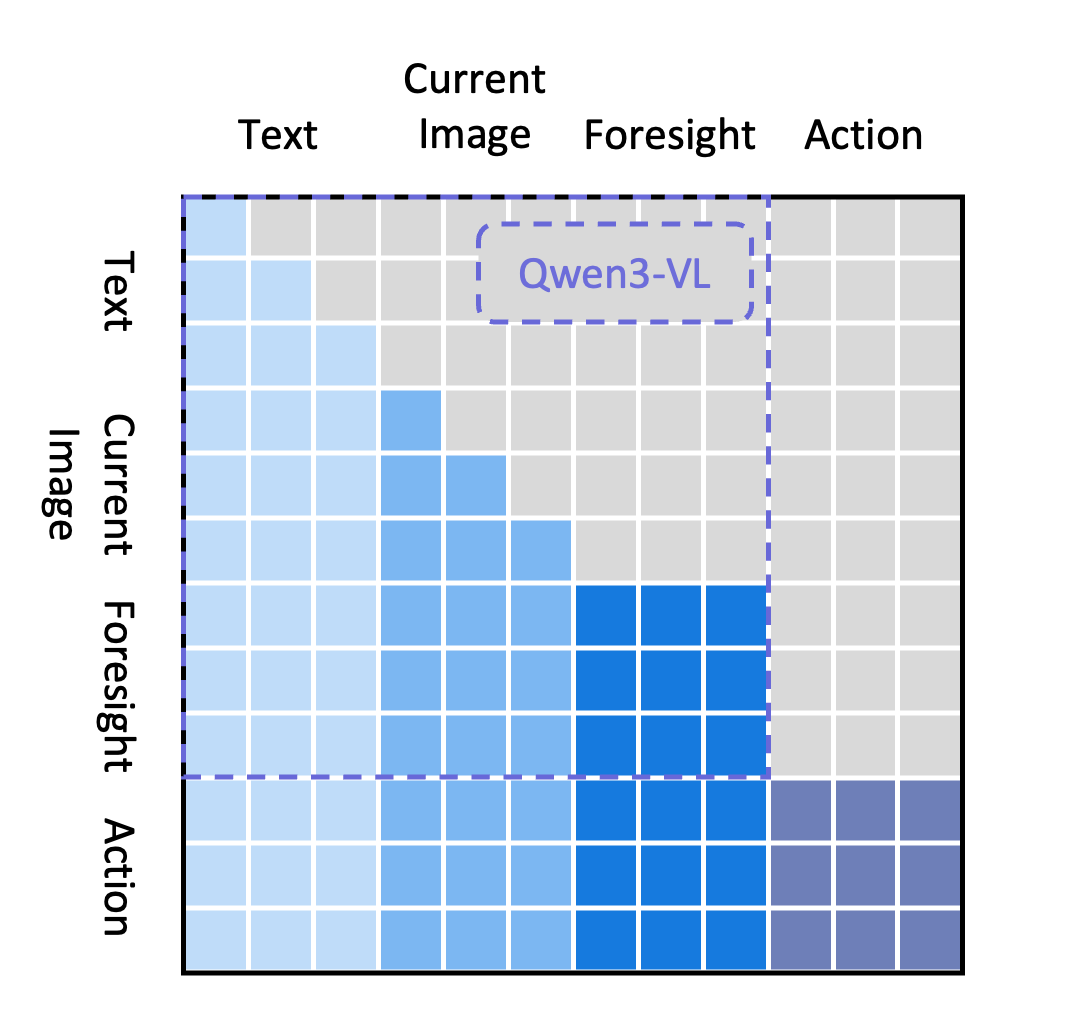}
    \caption{Attention mask design of FoMoVLA. }
    \label{fig:attention_mask}
\end{figure}

This asymmetric attention design ensures that (1)~instruction tokens provide unidirectional conditioning to all downstream modalities without being influenced by visual observations, (2)~foresight tokens aggregate global context through bidirectional self-attention to form a coherent future representation, and (3)~action tokens have full visibility over all available information. Fig.~\ref{fig:attention_mask} illustrates the attention mask structure.

\section{Additional Experimental Analysis}
\subsection{Inference Cost}

Tab.~\ref{tab:cost} compares the inference cost between the Vanilla baseline and FoMoVLA. The only additional overhead at deployment comes from processing the $K{=}16$ \texttt{<Foresight>} tokens appended to the input sequence. This increases median inference latency by only 9.4\,ms and allocated GPU memory by merely 0.1\,GB. These marginal costs are negligible compared to the substantial performance gains reported in the main paper, confirming that FoMoVLA introduces minimal deployment overhead.

\begin{table*}[t]
\centering
\setlength{\tabcolsep}{2pt}
\begin{tabular}{l *{5}{>{\centering\arraybackslash}m{2.1cm}}}
\toprule
\textbf{Task} & \textbf{Vanilla} & \textbf{+ Fut Pred.} & \textbf{+ Tracking} & \hfil\textbf{+ Fut Pred.}\hfil\newline\hfil\textbf{+ Tracking}\hfil & \hfil\textbf{+ Fut Pred.}\hfil\newline\hfil\textbf{+ Tracking} \hfil\newline\textbf{+ FCCA}\hfil \\
\midrule
PnPBottleToCabinetClose               & 46.0 & 64.0 & 70.0 & 68.0 & 72.0 \\
PnPCanToDrawerClose                    & 80.0 & 68.0 & 72.0 & 79.0 & 73.0 \\
PnPCupToDrawerClose                    & 54.0 & 52.0 & 50.0 & 51.0 & 41.0 \\
PnPMilkToMicrowaveClose                & 48.0 & 60.0 & 38.0 & 56.0 & 51.0 \\
PnPPotatoToMicrowaveClose              & 28.0 & 42.0 & 28.0 & 40.0 & 38.0 \\
PnPWineToCabinetClose                  & 46.0 & 44.0 & 56.0 & 54.0 & 49.0 \\
PnPNovelFromCuttingboardToBasket       & 48.0 & 48.0 & 52.0 & 57.0 & 54.0 \\
PnPNovelFromCuttingboardToCardboardbox & 40.0 & 54.0 & 36.0 & 52.0 & 53.0 \\
PnPNovelFromCuttingboardToPan          & 68.0 & 82.0 & 70.0 & 70.0 & 69.0 \\
PnPNovelFromCuttingboardToPot          & 52.0 & 62.0 & 42.0 & 61.0 & 67.0 \\
PnPNovelFromCuttingboardToTieredbasket & 56.0 & 62.0 & 44.0 & 51.0 & 55.0 \\
PnPNovelFromPlacematToBasket           & 42.0 & 60.0 & 54.0 & 55.0 & 60.0 \\
PnPNovelFromPlacematToBowl             & 44.0 & 54.0 & 60.0 & 56.0 & 51.0 \\
PnPNovelFromPlacematToPlate            & 48.0 & 54.0 & 74.0 & 57.0 & 71.0 \\
PnPNovelFromPlacematToTieredshelf      & 18.0 & 24.0 & 30.0 & 44.0 & 27.0 \\
PnPNovelFromPlateToBowl                & 60.0 & 62.0 & 70.0 & 58.0 & 51.0 \\
PnPNovelFromPlateToCardboardbox        & 50.0 & 54.0 & 60.0 & 53.0 & 61.0 \\
PnPNovelFromPlateToPan                 & 54.0 & 58.0 & 64.0 & 54.0 & 63.0 \\
PnPNovelFromPlateToPlate               & 70.0 & 74.0 & 76.0 & 76.0 & 76.0 \\
PnPNovelFromTrayToCardboardbox         & 38.0 & 40.0 & 62.0 & 56.0 & 60.0 \\
PnPNovelFromTrayToPlate                & 56.0 & 58.0 & 78.0 & 74.0 & 78.0 \\
PnPNovelFromTrayToPot                  & 50.0 & 68.0 & 64.0 & 53.0 & 71.0 \\
PnPNovelFromTrayToTieredbasket         & 36.0 & 44.0 & 50.0 & 51.0 & 45.0 \\
PnPNovelFromTrayToTieredshelf          & 16.0 & 18.0 & 34.0 & 32.0 & 30.0 \\
\midrule
\textbf{Average}                       & 47.8 & 54.4 & 55.6 & 56.6 & \textbf{56.9} \\
\bottomrule
\end{tabular}%
\caption{Ablation study on RoboCasa GR-1 Tabletop: per-task success rate (\%) across 24 tasks (50 rollouts/task). Each column progressively adds a component to the Vanilla baseline.}
\label{tab:robocasa_ablation}
\end{table*}

\begin{figure}[ht]
    \centering
    \includegraphics[width=\columnwidth]{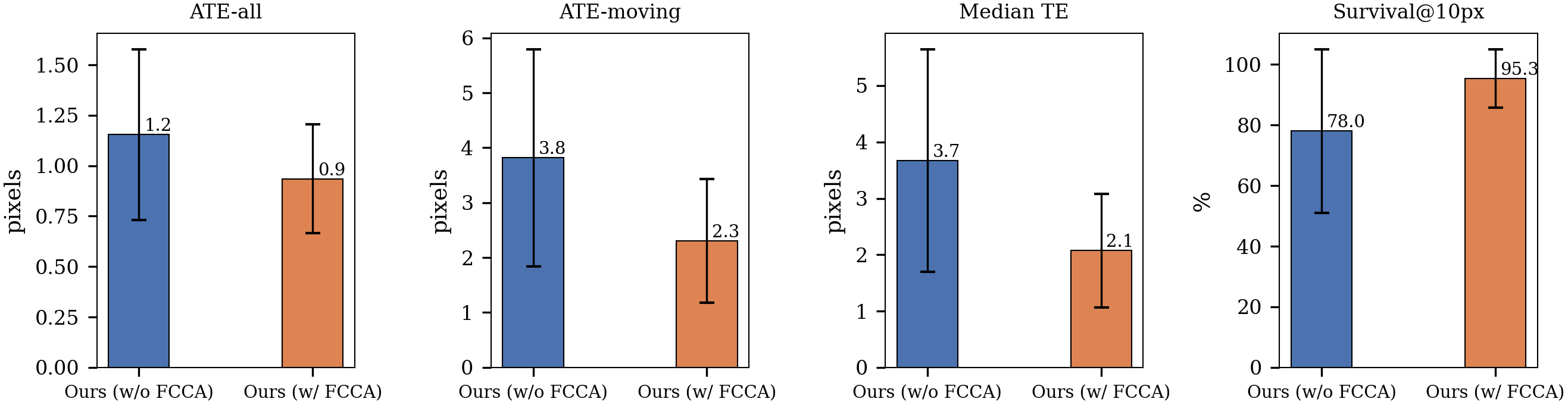}
    \caption{Quantitative comparison of point tracking accuracy with and without FCCA. Incorporating future-conditioned cross-attention reduces trajectory errors (ATE-all, ATE-moving, Median TE) and substantially improves Survival@10px, confirming that future-state conditioning provides a strong geometric prior for motion prediction.}
    \label{fig:fcca_bar}
\end{figure}

\begin{figure*}[ht]
    \centering
    \includegraphics[width=0.9\textwidth]{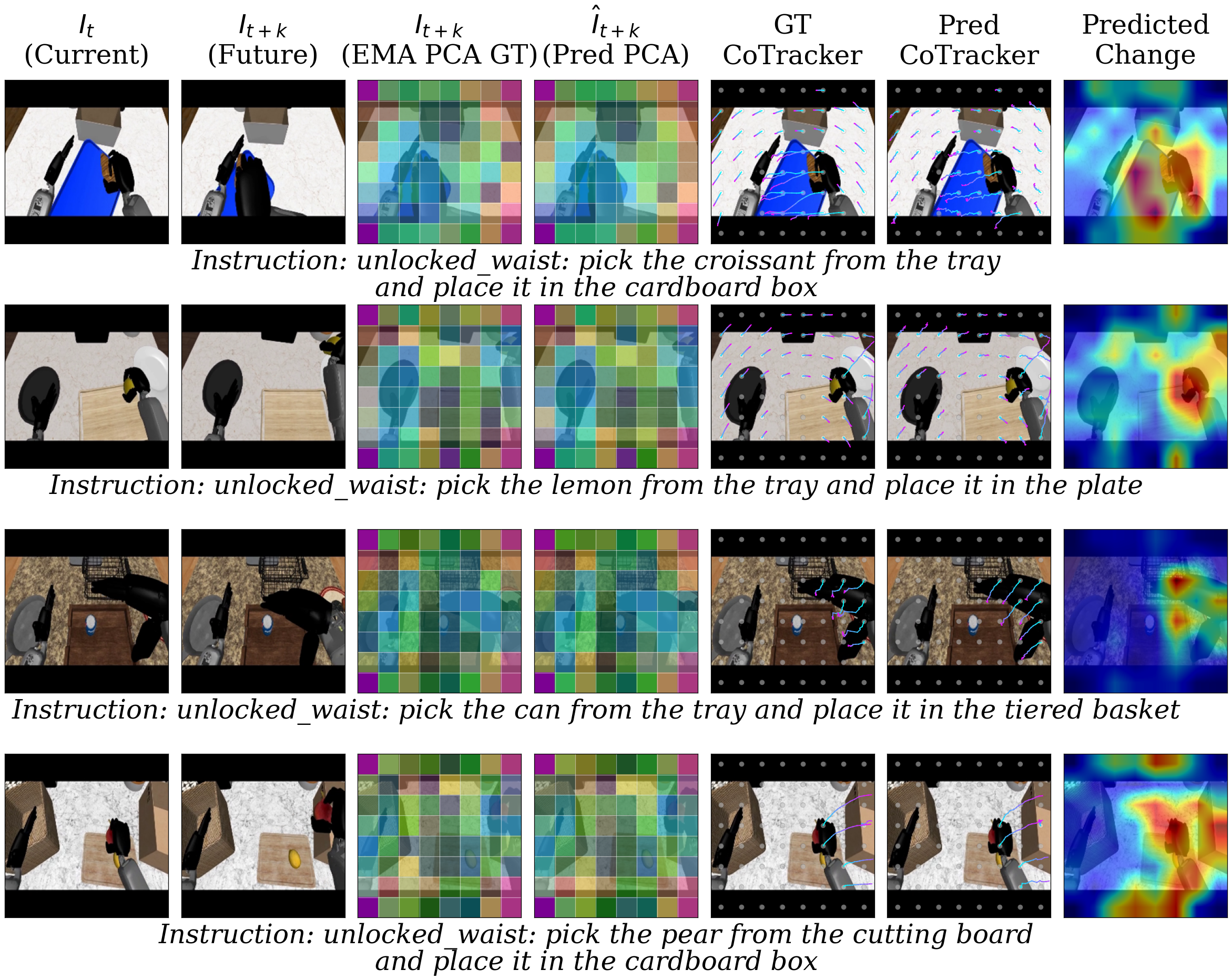}
    \caption{Qualitative results on RoboCasa GR-1 Tabletop (egocentric view). From left to right: current observation, GT future frame, GT future PCA features, predicted foresight PCA, GT point tracks, predicted point tracks, and predicted change heatmap. The model successfully handles non-stationary egocentric viewpoints, producing accurate future predictions and motion trajectories.}
    \label{fig:robocasa_vis}
\end{figure*}

\begin{table}[ht]
\centering
\small
\begin{tabular}{lcc}
\toprule
\textbf{Metric} & \textbf{Vanilla} & \textbf{FoMoVLA} \\
\midrule
Median Latency (ms)        & 94.3   & 103.7 \\
Mean Latency (ms)          & 97.5   & 103.8 \\
Allocated GPU Memory (GB)  & 9.3    & 9.4 \\
\bottomrule
\end{tabular}
\caption{Inference cost comparison. Latency and memory are measured on a single H20 GPU with batch size 1, averaged over 1000 runs after 50 warmup iterations.}
\label{tab:cost}
\end{table}

\subsection{Quantitative Effect of FCCA on Point Tracking}
To quantify the effect of FCCA on point tracking quality, we evaluate models trained with and without FCCA on 10,000 samples from the LIBERO-10 training set using a frozen CoTracker teacher as ground truth. We report four metrics: (1)~ATE-all, the average trajectory error over all 64 tracked grid points; (2)~ATE-moving, restricted to points whose ground-truth displacement exceeds 3\,px (capturing task-relevant motion); (3)~Median TE, the median trajectory error over moving points (robust to outliers); and (4)~Survival@10px, the fraction of moving points whose final-frame error remains below 10\,px.

As shown in Fig.~\ref{fig:fcca_bar}, incorporating FCCA yields consistent improvements across all metrics. ATE-all decreases from 1.2\,px to 0.9\,px ($-$25.0\%), and ATE-moving drops from 3.8\,px to 2.3\,px ($-$39.5\%), indicating substantially more accurate displacement predictions for task-relevant moving points. Median TE similarly decreases from 3.7\,px to 2.1\,px ($-$43.2\%), confirming that the improvement is not driven by outlier correction alone. Most notably, Survival@10px increases from 78.0\% to 95.3\% ($+$17.3\,pp), demonstrating that FCCA enables the vast majority of tracked points to converge within a tight error bound by the end of the action chunk. These results confirm that explicitly conditioning motion prediction on the predicted future state---rather than treating future prediction and point tracking as independent objectives---yields significantly more accurate and reliable trajectory estimation.

\subsection{RoboCasa GR-1 Tabletop Ablation}

\subsubsection{Quantitative Results}

Tab.~\ref{tab:robocasa_ablation} presents a component-wise ablation on the RoboCasa GR-1 Tabletop benchmark (24 tasks, 50 rollouts per task). Starting from the Vanilla baseline (47.8\%), adding future prediction alone improves the average success rate to 54.4\% (+6.6\%), and adding point tracking alone yields 55.6\% (+7.8\%), indicating that both auxiliary objectives independently enhance policy learning. Combining the two objectives further boosts performance to 56.6\% (+8.8\%), demonstrating their complementarity: future prediction provides semantic goal-state understanding while tracking supplies geometric motion reasoning. Finally, incorporating FCCA achieves the best overall performance of 56.9\% (+9.1\%), confirming that explicit conditioning of motion prediction on future representations is essential for full synergy. These trends are consistent with the LIBERO ablation (Tab.~\ref{tab:main}), validating that the proposed components generalize across different robot embodiments and camera configurations (egocentric vs.\ third-person).

\subsubsection{Qualitative Results}
\label{sec:appendix_robocasa_vis}

To further validate the generality of FoMoVLA beyond fixed-camera setups, we present qualitative visualizations on the RoboCasa GR-1 Tabletop benchmark (Fig.~\ref{fig:robocasa_vis}). Unlike LIBERO, which uses a static third-person camera, RoboCasa provides egocentric observations mounted on the robot's head, introducing continuous viewpoint changes induced by the robot's body motion. This setting requires the model to disentangle ego-motion from task-relevant object motion under a non-stationary camera.

Overall, FoMoVLA produces coherent predictions: the predicted foresight closely matches ground-truth future features under viewpoint shifts, and the predicted point trajectories correctly capture both ego-motion (global camera displacement) and local arm manipulation motion. The predicted heatmap change further confirms that the model concentrates predictive attention on task-relevant regions (target object and end-effector path) rather than static background, demonstrating spatially grounded foresight under dynamic egocentric viewpoints.

\end{document}